\documentclass[lettersize,journal]{IEEEtran}
\usepackage{amsmath,amsfonts}
\usepackage{algorithmic}
\usepackage{algorithm}
\usepackage{array}
\usepackage[caption=false,font=normalsize,labelfont=sf,textfont=sf]{subfig}
\usepackage{textcomp}
\usepackage{stfloats}
\usepackage{url}
\usepackage{verbatim}
\usepackage{graphicx}
\usepackage{cite}
\usepackage{multirow}
\usepackage{amssymb}
\usepackage{url}
\usepackage{amsmath}
\usepackage{graphicx}
\usepackage[printonlyused]{acronym}
\usepackage{tabularx}
\usepackage{multirow}
\usepackage{booktabs}
\usepackage{graphics}
\usepackage{multicol}
\usepackage{subcaption}
\usepackage{float}
\usepackage{xcolor}
\usepackage{colortbl}

\newcolumntype{Y}{>{\centering\arraybackslash}X}

\begin{document}

\title{An Efficient and Effective Encoder Model for Vision and Language Tasks in the Remote Sensing Domain}

\author{João Daniel Silva,
        João Magalhães,
        Devis Tuia,~\IEEEmembership{Fellow,~IEEE,}
        Bruno Martins,~\IEEEmembership{Senior Member,~IEEE}
\thanks{João Daniel Silva and Bruno Martins are with INESC-ID, Instituto Superior Técnico, University of Lisbon, Portugal (e-mail: joao.daniel.silva@tecnico.ulisboa.pt).}
\thanks{João Magalhães is with the Department of Computer Science, Faculty of Science and Technology, Universidade NOVA de Lisboa, Portugal.}
\thanks{Devis Tuia is with the School of Architecture, Civil and Environmental Engineering, EPFL, 1950 Sion, Switzerland}
\thanks{Manuscript received -, -; revised -, -}}

\markboth{Journal of \LaTeX\ Class Files,~Vol.~14, No.~8, August~2021}%
{Shell \MakeLowercase{\textit{et al.}}: A Sample Article Using IEEEtran.cls for IEEE Journals}


\maketitle

\begin{abstract}
The remote sensing community has recently seen the emergence of methods based on Large Vision and Language Models (LVLMs) that can address multiple tasks at the intersection of computer vision and natural language processing. To fully exploit the potential of such models, a significant focus has been given to the collection of large amounts of training data that cover multiple remote sensing-specific tasks, such as image captioning or visual question answering. However, the cost of using and training LVLMs is high, due to the large number of parameters. While multiple parameter-efficient adaptation techniques have been explored, the computational costs of training and inference with these models can remain prohibitive for most institutions. In this work, we explore the use of encoder-only architectures and propose a model that can effectively address multi-task learning while remaining compact in terms of the number of parameters. In particular, our model tackles combinations of tasks that are not typically explored in a unified model: the generation of text from remote sensing images and cross-modal retrieval. The results of our \textbf{GeoMELT} model - named from Multi-task Efficient Learning Transformer - in established benchmarks confirm the efficacy and efficiency of the proposed approach.
\end{abstract}


\section{Introduction}

\begin{figure}[t]
  \centering
   \includegraphics[width=\linewidth]{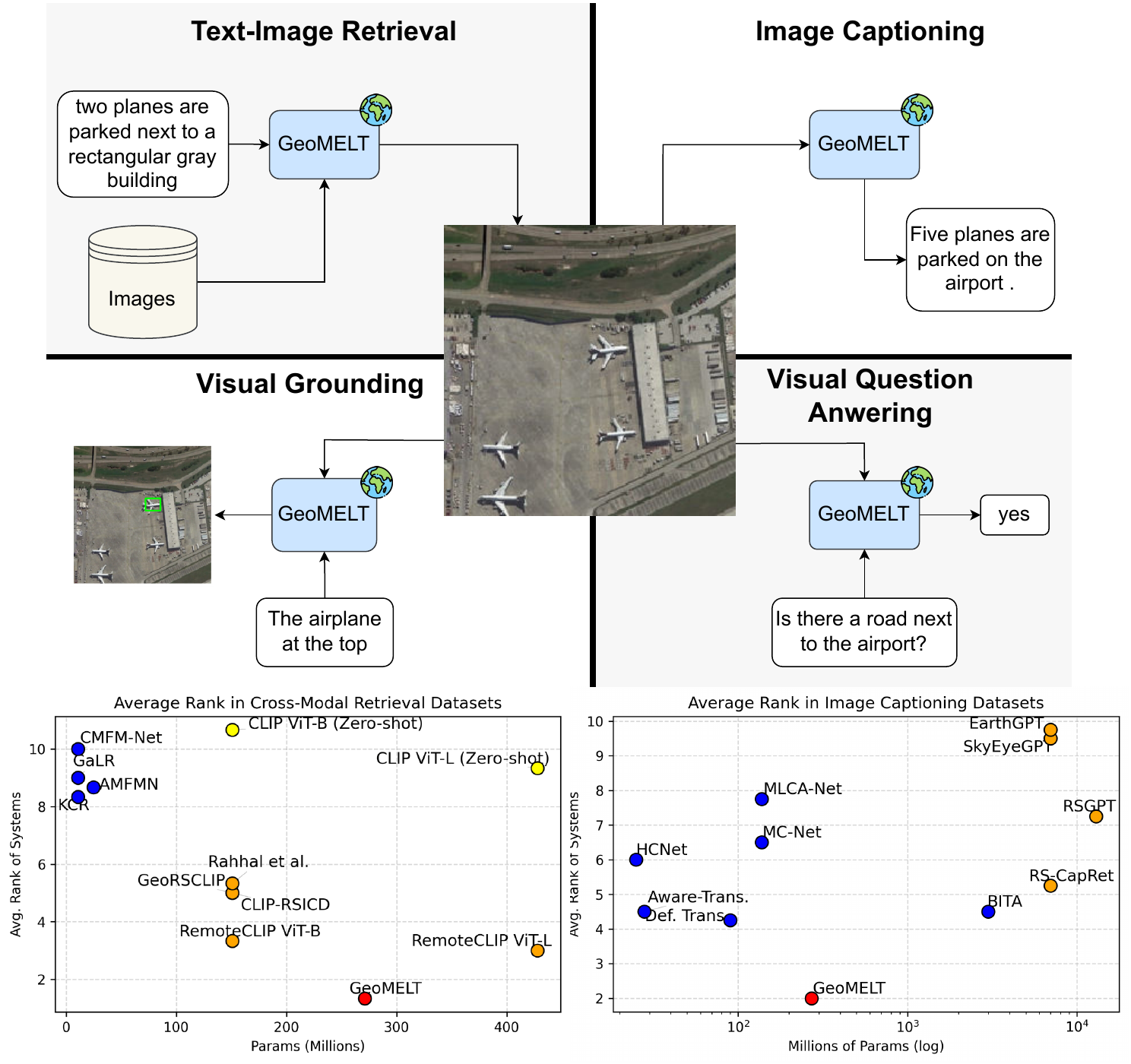}

   \caption{GeoMELT is a multi-task vision and language model for remote sensing. Across image captioning and cross-modal retrieval evaluatuions, each aggregated over multiple datasets, GeoMELT consistently achieves top overall results while staying efficient regarding the number of parameters.
   }
   \label{fig:performance}
\end{figure}

\IEEEPARstart{T}{he} intersection of computer vision and natural language processing has been the subject of significant research~\cite{liu2023visual,laurenccon2024building,wang2024cogvlm,radford2021Learninga}, and the remote sensing community is no exception~\cite{tuia2024artificial,li2024vision,muhtar2024lhrs,kuckreja2024geochat}. Multiple approaches specific to the remote sensing domain were proposed, and evaluation benchmarks were systematized for tasks such as image captioning~\cite{cheng2022NWPUCaptions,lu2017exploring,zhao2021Highresolution}, text-image retrieval~\cite{lu2017exploring,yuan2022exploring,mi2024knowledge}, visual question answering~\cite{lobry2020RSVQA,chappuis2022prompt,yuan2022easy,zhang2023spatial}, and visual grounding~\cite{zhan2023rsvg,sun2022visual}. From a practical point of view, the development of these methods can help a wider goal of enabling non-expert users to interact with Earth Observation data through natural language queries instead of traditional mapping, with potential benefits regarding education and aid~\cite{sarkar2021vqa-aid,rahnemoonfar2021floodnet}.

At first, the methods proposed in previous work relied on models specialized in each of these tasks, and were limited by both the capacity of the models and the available data to train them. Generative tasks were based on encoder-decoder architectures~\cite{cheng2022NWPUCaptions,li2020multi}, with a particular focus on attention mechanisms that attended to the details of remote sensing images, while approaches to align embeddings from images and text were developed for text-image retrieval~\cite{yuan2022remote,mi2022knowledge}, and open vocabulary concept exploration~\cite{zermatten2025learning,marnoto2025generalized}.

A particular task at the intersection of vision and language that has been widely studied in remote sensing is cross-modal retrieval, where a query based on an image (or text) is matched with relevant items corresponding to text descriptions (or images). This task is typically addressed with encoder-based models trained with different strategies that align the embeddings of each modality~\cite{yuan2022remote,remoteclip,silva2024multilingual}. Despite the good results, previous models are typically constrained to this task, and the inclusion of other tasks, e.g., involving text generation, remains difficult and underexplored~\cite{silva2024large}.

Mirroring advances in computer vision~\cite{alayrac2022flamingo,radford2021Learninga}, the community has more recently seen the emergence of methods that scaled both the number of parameters and the volume of training data~\cite{remoteclip,kuckreja2024geochat,li2024vision}. Leveraging the existing high volume of unlabeled image repositories, along with self-supervised objectives, multiple vision encoders were proposed that can be used as backbones for downstream tasks~\cite{cong2022satmae,sun2022Ringmo,wang2022Advancing,sumbuel2025smarties,rolf2021generalizable}, and which hold the promise of providing feature representations transferable across remote sensing problems. Meanwhile, the introduction of Large Language Models (LLMs)~\cite{brown2020language,touvron2023llama2,groeneveld2024olmo} led to breakthroughs in natural language processing tasks and text generation quality, opening the way for principled human-machine interaction through natural language queries. The breakthroughs associated to LLMs have motivated research on how to incorporate other modalities beyond text, in particular images, such that requests in natural language can be made about visual contents, resulting in Large Vision and Language Models (LVLMs) such as Flamingo~\cite{alayrac2022flamingo}, LLaVA~\cite{liu2023visual}, or Qwen-VL~\cite{Qwen2.5-VL}. These developments have also influenced the remote sensing field, which has seen an emergence of multiple contributions, including GeoChat~\cite{kuckreja2024geochat}, LHRS-Bot~\cite{muhtar2024lhrs}, or SkyEyeGPT~\cite{zhan2025skyeyegpt}. However, a particular difficulty is that although these LVLMs have been shown to perform well in benchmarks regarding tasks such as image captioning, Visual Question Answering (VQA), or Visual Grounding (VG), they are computationaly very demanding, with the costs of inference and tuning, in particular, remaining very high. Deploying such models on edge devices like aerial vehicles remains challenging~\cite{zhang2025satellite}, although this would be a natural application for these models. While multiple parameter-efficient adaptation techniques have been introduced~\cite{hu2022lora,yuan2023parameter}, it remains important to explore other compact architectures that can achieve good results in such scenarios, to reduce computational costs and make those models accessible to everyone willing to customize and tune them on more downstream tasks.

This work presents GeoMELT, i.e. a lightweight vision-and-language model designed for a wide range of tasks with remote sensing imagery, with only \textbf{271M} parameters and depicted in Fig.~\ref{fig:performance}. We leverage a multimodal encoder architecture trained with an objective that balances text generation and embedding alignment. Specifically, GeoMELT can be used as a dual encoder for text-image retrieval, and as a cross-encoder for generating text conditioned on an image. The text generation capabilities of GeoMELT are not limited to image captioning but also include visual grounding and visual question answering. As a result, GeoMELT efficiently addresses multiple tasks without compromising performance, achieving state-of-the-art results across different benchmarks.

The rest of the paper is organized as follows: Section~\ref{rw} covers related work, while Section~\ref{method} details the proposed method. Section~\ref{experimental} describes the experimental setup, and Section~\ref{results} discusses the obtained results. Finally, Section~\ref{conclusion} describes the main conclusion and future research avenues.

\section{Related Work}\label{rw}

The increasing volume of Earth Observation data, combined with advances in natural language processing and computer vision, has led to multiple vision-language proposals tailored to the remote sensing domain. This section surveys such efforts.

\subsection{Large Vision and Language Models in Remote Sensing}\label{section:lvlms}

The success of Large Language Models (LLMs) has motivated research into the integration of vision encoders, typically from CLIP models, with large Transformer decoders for language generation, resulting in Large Vision and Language Models (LVLMs)~\cite{alayrac2022flamingo,liu2023visual,chen2023minigpt,wang2024cogvlm}. Regarding remote sensing–specific methods, an early work is VLCA~\cite{wei2023VLCA}, which combined the CLIP image encoder with GPT-2~\cite{radford2019Language}. Pioneering the use of more recent LLMs, RSGPT~\cite{hu2025rsgpt} is a fine-tuned InstructBLIP model~\cite{dai2023instructblip} over a manually curated captioning dataset. Multiple multi-task LVLMs soon followed, namely LHRS-Bot~\cite{muhtar2024lhrs}, SkyEyeGPT~\cite{zhan2025skyeyegpt}, GeoChat~\cite{kuckreja2024geochat}, or VHM~\cite{pang2025vhm}, that address tasks such as image captioning, visual question answering, and visual grounding. A significant effort has been made toward developing data mixtures to train these models effectively across various remote sensing tasks. However, the cost of training these models is very high, as just the LLM component has frequently more than 7 billion parameters. This has led researchers to apply parameter-efficient techniques, such as LoRA~\cite{hu2022lora}, or to only update a subset of the total parameters of these models, particularly the image encoder and the multimodal connector. Another point is that despite the breadth of tasks that these models can perform, cross-modal retrieval remains difficult and underexplored due to the different nature of the text generation and alignment of embeddings tasks~\cite{silva2024large}. A compact and lightweight model that can address both text generation and cross-modal retrieval remains to be proposed in the remote sensing domain.

\subsection{Generating Text Conditioned on Remote Sensing Images}

Several tasks involve generating text conditioned on the visual contents of an image. Among these, image captioning is a widely studied task, with early approaches based on the encoder-decoder framework~\cite{ramos2022Using,zhao2021Highresolution,cheng2022NWPUCaptions,li2021truncation}, while subsequent work focused on specialized attention mechanisms that can capture fine-grained and multi-scale details~\cite{yuan2020exploring,cheng2022NWPUCaptions,li2020multi}. For example, MLCA-Net employs a multi-level attention and channel attention mechanisms to better focus on objects~\cite{cheng2022NWPUCaptions}. Methods based on the Transformer decoder include the Aware-Transformer, which employs spatial and channel attention mechanisms~\cite{cao2023aware}, and HCNet~\cite{yang2024hcnet}, which extracts hierarchical features from the input image by introducing a cross-modal feature interaction module to fuse the image features with the text features during generation. While these methods can effectively output descriptions of remote sensing images, the image attention mechanisms are increasingly complex, making it difficult to leverage the recent advances in vision and language models discussed in Section~\ref{section:lvlms}. 

Visual Question Answering (VQA) is also another important task in the remote sensing community~\cite{lobry2020RSVQA,rahnemoonfar2021floodnet,chappuis2022prompt}. However, many approaches have treated VQA as a classification problem, and only with the advent of LVLMs has a generative approach been followed~\cite{hu2025rsgpt,zhan2025skyeyegpt}. Earlier works leveraged different types of vision and text backbones with multimodal fusion strategies~\cite{chappuis2022PromptRSVQA,bazi2022bi,yuan2022easy}. Similarly, Visual Grounding (VG) also began as numerical coordinate prediction~\cite{sun2022visual,zhan2023rsvg,marnoto2025generalized,chen2025rsrefseg} and only recently was addressed in a generative approach, with LVLMs generating the coordinates of the object of interest as text tokens~\cite{kuckreja2024geochat,muhtar2024lhrs}. Before that, approaches like GeoVG~\cite{sun2022visual} used a language
encoder to model geospatial relations in the text request, an adaptive region attention, and a fusion module, while RSVG~\cite{zhan2023rsvg} leveraged ViT and BERT to obtain input features, and a multi-scale and multimodal feature aggregation mechanism for bounding box prediction. 

\subsection{Cross-modal Retrieval in Remote Sensing}

Cross-modal retrieval in remote sensing focuses on aligning visual and textual representations to enable image-to-text and text-to-image search. The literature in the area features several specialized approaches aiming to build representations from each modality that align in a common embedding space~\cite{yuan2022remote,mi2022knowledge,remoteclip,yuan2021lightweight}. GaLR~\cite{yuan2022remote} introduced a mechanism to combine global attention and local features obtained with a graph convolutional network. To address the gap in information between text descriptions and the image contents, KCR~\cite{mi2022knowledge} and KTIR~\cite{mi2024knowledge} leverage a knowledge graph to enrich the text descriptions. Hoxha et al.~\cite{hoxha2020toward} proposed to first generate a description for the image and retrieve captions based on the word similarity with the candidates. AMFMN~\cite{yuan2022exploring} employs an attention mechanism over multiscale features.

One particular line of research that has been very successful relates to the adaptation of CLIP~\cite{radford2021Learninga} to the remote sensing domain. Yuan et al.~\cite{yuan2023parameter} developed a CLIP adapter specifically for the remote sensing image-text retrieval task. RemoteCLIP~\cite{remoteclip} aggregated datasets from different tasks to generate image-caption pairs, resulting in a strong performance. GeoRSCLIP~\cite{zhang2024rs5m} collected image-caption pairs from publicly available datasets from computer vision that also contained remote sensing images, and developed a pipeline to generate synthetic captions up to 5M samples. RS-M-CLIP~\cite{silva2024multilingual} combined the CLIP and DINO~\cite{caron2021emerging} training objectives, together with multilingual data augmentation.

\section{Learning to Generate Text and Align Image and Text Embeddings}\label{method}

\begin{figure*}[t]
  \centering
   \includegraphics[width=\linewidth]{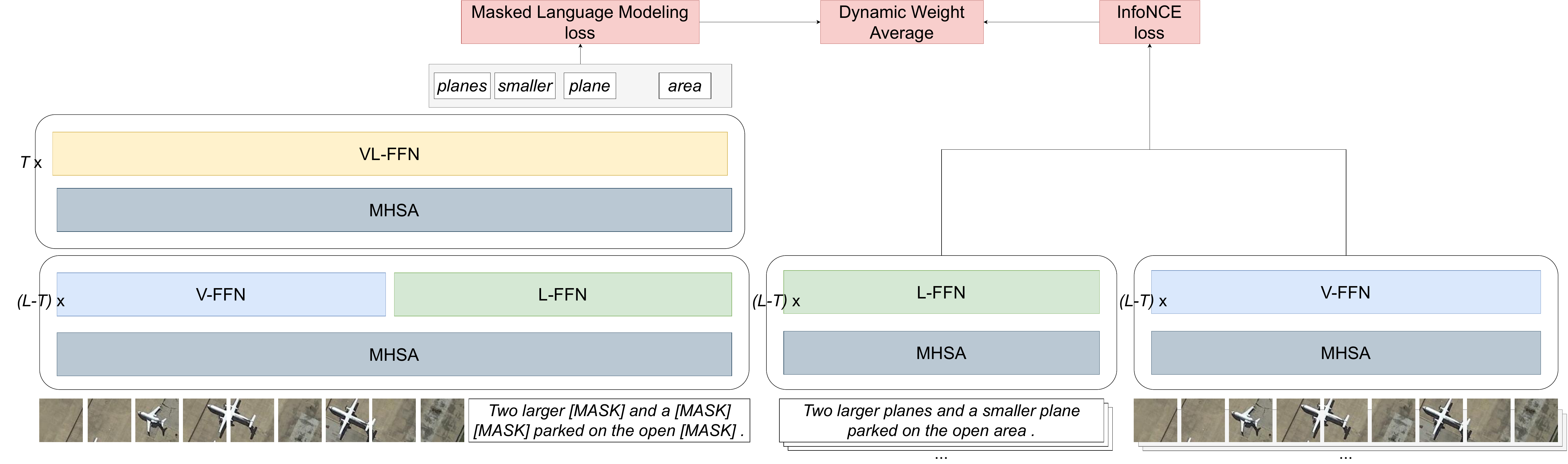}

   \caption{Overview of the GeoMELT model based on a multimodal encoder architecture with a total of 271M parameters. The earlier layers contain a shared Multi-Head Self-Attention (MHSA) layer and Feed-Forward Networks (FFN) that are specific to each modality. The top layers contain a shared FFN. Masked language modeling is used for text generation training. Importantly, text generation with GeoMELT can go beyond image captioning and include visual grounding and visual question answering tasks. For retrieval tasks, the model can be used as a dual encoder and trained with contrastive learning.}
   \label{fig:method}
\end{figure*}

Our work is motivated by recent developments in computer vision and natural language processing that have made possible the development of LVLMs for the remote sensing domain, which can simultaneously address multiple tasks. However, in this work, we depart from this line of research due to the significant computational constraints in terms of inference and finetuning that LVLMs impose. Additionally, our objective is to develop a method that is lightweight and capable of handling tasks that are rarely addressed together, specifically cross-modal retrieval, which aligns representations across different modalities, and text generation conditioned on images. To achieve this, we adopt a different architectural choice corresponding to a multimodal encoder.

In particular, GeoMELT leverages a unified multimodal encoder based on BEiT-3~\cite{wang2023image}. This architecture considers modality-specific encoding and cross-modality fusion, allowing for a compact method, with parameters shared across modalities, involving a total of 271M parameters.

The multimodal encoder consists of Multiway Transformer (MWT) blocks~\cite{bao2022vlmo}, each containing a shared Multi-Head Self-Attention mechanism (MHSA) followed by multiple Feed-Forward Networks (FFNs). These FFNs act as modality-specific experts, specialized for language, vision, or joint vision–language representations. Each input token is routed to an expert based on the corresponding modality. We follow BEiT-3 with a vision expert and a language expert at each layer, with the top-3 layers also containing vision-language experts. This combination of shared parameters, namely the self-attention and modality-specific parameters, allows the alignment between modalities while enabling the modeling of modality-specific features. The modularity of the architecture allows for different types of use, namely as (a) a vision encoder-only backbone for image-only tasks, such as image classification, (b) a dual encoder for text-image retrieval, or (c) a cross-encoder for multimodal representation or generation tasks.

GeoMELT addresses two main tasks, namely text generation conditioned on an image and text-image retrieval. The model is trained over a dataset of $M$ pairs, each containing a remote sensing image $\mathbf{x}_i$ and a textual description $\mathbf{y}_i$: $\mathcal{D}=\{\mathbf{x}_{i}, \mathbf{y}_i \}_{i=1}^{L}$. The textual description $\mathbf{y}_i$ can vary across different possible tasks. It can be a caption for image captioning, a question and the corresponding answer for VQA, or a region description together with a set of image coordinates for visual grounding. The following is a description of our training procedure, and a graphical depiction is given in Fig.~\ref{fig:method}.

\subsection{Text Generation}

The first main task is text generation, which includes image captioning, visual grounding, and visual question answering, each requiring the model to produce coherent textual outputs conditioned on visual inputs. Given that GeoMELT is encoder-based, text generation is performed as conditional generation over the \texttt{[MASK]} tokens provided with the text. Specifically, during training, a sequence of text tokens is given to the model, where a percentage of these are masked with probability $p_M$. The model is then trained to recover those tokens, as in the Masked Language Modeling (MLM) objective. Notice nonetheless, that when predicting text tokens, the model should not attend to future tokens, as the prediction should only be based on the image contents and previous text tokens. This is achieved by applying an attention mask that allows bidirectional attention of image patches over other image patches, and at the same time restricting text tokens to only attend to the image tokens, the previous text tokens, and themselves. During inference, the \texttt{[MASK]} token is appended at the end of the sequence that is being generated and, at each time step $t$, a new token is predicted in an autoregressive manner.

Formally, given an original text sequence $\mathbf{y} = (y_1, \dots, y_n)$ for model training another sequence $\tilde{\mathbf{y}}$ is obtained that contains the original sequence together with the tokens masked at randomly selected positions $\mathcal{M} \subset \{1, \dots, n\}$. The original tokens are recovered through the cross-entropy loss by the model $P_\theta$:

\begin{equation}
    \mathcal{L}_{\mathrm{MLM}} = - \sum_{i \in \mathcal{M}} \log P_\theta \big( y_i \mid \tilde{\mathbf{y}} \big).
\end{equation}\label{eq:mlm}

Regarding masking strategies, for image captioning the \texttt{[MASK]} tokens can be placed at any position in the description $y$. For VQA, we have observed that masking only the answer tokens hinders model learning, whereas masking both the question and the answer leads to better performance. We therefore also allow the \texttt{[MASK]} tokens to appear at any position in this case. For the visual grounding task, we treat it as a particular case of text generation, where the query of the object description $y_i$ is associated with the bounding-box coordinates $(\langle x_1 \rangle, \langle y1\rangle, \langle x2\rangle, \langle y2 \rangle)$, where $(\langle x_1 \rangle, \langle y1\rangle)$ and $(\langle x_2 \rangle, \langle y2\rangle)$ represent the coordinates of the top-left and bottom-right corners of the object's bounding box, respectively. The coordinate values are normalized relative to the image dimensions, scaling them to the range of $[0,100]$. This constrains the range of possible values to be generated and allows the processing of images at different resolutions. During training, each coordinate is replaced with the token \texttt{[MASK]} and the model has to predict the correct value, using the same loss as described previously.

Moreover, to better distinguish between tasks, a short prompt is appended to the beginning of the text, with short and long captioning being associated with "\texttt{short caption:}" and "\texttt{long caption:}", respectively, VQA with "\texttt{vqa:}" and VG with "\texttt{bounding-box:}". 

\subsection{Cross-Modal Retrieval}

\begin{table}[t!]
\centering
\caption{Overview of the datasets used to train GeoMELT, along with a comparison to other datasets for training remote sensing vision–language models. “Samples” indicates the number of training and test instances used from each dataset.}
\label{tab:datasets}
\resizebox{\linewidth}{!}{
\begin{tabular}{llcc}
\toprule
Task                                       & Dataset       & \# Train Samples & \# Test Samples \\ \midrule
\multirow{5}{*}{Image Captioning}          & NWPU-Captions~\cite{cheng2022NWPUCaptions} & 126,000          & 15,750          \\
                                           & RSICD~\cite{lu2017exploring}         & 43,670           & 5,465           \\
                                           & UCM-Captions~\cite{qu2016deep}           & 8,400            & 1,050           \\
                                           & Sydney-Captions~\cite{qu2016deep}        & 2,485            & 290             \\
                                           & RSITMD~\cite{yuan2022exploring}        & 21,455           & 2,260           \\ \midrule
\multirow{2}{*}{Detailed Image Captioning}           & RSICap~\cite{hu2025rsgpt}        & 2,585            & 100             \\
                                           & VRSBench~\cite{li2024vrsbench}      & 20,264           & 9,350           \\ \midrule
\multirow{2}{*}{Visual Question Answering} & RSVQA-HR~\cite{lobry2020RSVQA} & 20,000           & 131,468 \\
                                           & RSVQA-LR~\cite{lobry2020RSVQA} & 10,000           & 7,057\\ \midrule
Visual Grounding                            & DIOR-RSVG~\cite{zhan2023rsvg} & 26,991           & 7,500           \\ 
 \toprule
\multicolumn{4}{c}{Comparison in the Volume of Training Data}                                                                \\ 
\midrule
Method                                    & \multicolumn{3}{c}{\# Train Samples}               \\ \midrule
RemoteCLIP~\cite{remoteclip}& \multicolumn{3}{c}{829k}  \\
RS5M~\cite{zhang2024rs5m} & \multicolumn{3}{c}{5M}  \\
SkyeEyeGPT~\cite{zhan2025skyeyegpt} & \multicolumn{3}{c}{968k}                               \\
LHRS-Bot~\cite{muhtar2024lhrs} & \multicolumn{3}{c}{1.15M+}                               \\
EarthGPT - MMRS-1M~\cite{zhang2024earthgpt} & \multicolumn{3}{c}{1.1M}                               \\
\cellcolor{red!15}Ours & \multicolumn{3}{c}{\cellcolor{red!15}\textbf{282k}}               \\
\bottomrule
\end{tabular}
}
\end{table}

\begin{table*}[t!]
\centering
\caption{Image-text retrieval results in the remote sensing domain. Results obtained from our own experiments, rather than collected from prior publications, are marked with~\dag. MWT denotes Multiway Transformers~\cite{bao2022vlmo}. Reported total parameters are in millions, where * indicates counts based solely on the vision backbone.
}
\label{table:clip-objective}
\resizebox{\textwidth}{!}{
\begin{tabular}{clcccccccccc}
\toprule
\multirow{2}{*}{Dataset} & \multicolumn{1}{c}{\multirow{2}{*}{Method}} & \multirow{2}{*}{\shortstack{\\Vision\\Backbone}} & \multirow{2}{*}{\shortstack{\\Finetuning\\Data}} & \multirow{2}{*}{\shortstack{\\Total\\Params}} & \multicolumn{3}{c}{Image-Text Retrieval} & \multicolumn{3}{c}{Text-Image Retrieval} & \multirow{2}{*}{\begin{tabular}[c]{@{}c@{}}Mean\\ Recall\end{tabular}} \\ \cmidrule{6-11}
 & \multicolumn{1}{c}{} &  &  & & R@1 & R@5 & R@10 & R@1 & R@5 & R@10 &  \\ \midrule
{\multirow{12}{*}{\rotatebox{90}{RSICD}}} & AMFMN~\cite{hoxha2020toward} & ResNet50 & RSICD & 25* & 5.39 & 15.08 & 23.40 & 4.90 & 18.28 & 31.44 & 16.42 \\
\multicolumn{1}{l}{} & CMFM-Net~\cite{yu2022text} & ResNet18 & RSICD & 11* & 5.40 & 18.66 & 28.55 & 5.31 & 18.57 & 30.03 & 17.75 \\
\multicolumn{1}{l}{} & GaLR~\cite{yuan2022remote} & ResNet18 & RSICD & 11* & 6.59 & 19.85 & 31.04 & 4.69 & 19.48 & 32.13 & 18.96 \\
\multicolumn{1}{l}{} & KCR~\cite{mi2022knowledge} & ResNet18 & RSICD & 11* & 4.76 & 18.59 & 27.20 & 5.84 & 22.31 & 36.12 & 19.14 \\
\multicolumn{1}{l}{} & CLIP~\cite{radford2021Learninga}\dag & ViT-B & Zero-shot & 151 & 4.58 & 14.55 & 23.70 & 5.80 & 16.85 & 28.23 & 15.62 \\
\multicolumn{1}{l}{} & CLIP~\cite{radford2021Learninga}\dag & ViT-L & Zero-shot & 428 & 6.04 & 17.48 & 27.54 & 5.03 & 19.03 & 30.25 & 17.56 \\
\multicolumn{1}{l}{} & Rahhal et al.~\cite{rahhal2022multilanguage} & ViT-B & RSICD & 151 & 10.70 & 29.64 & 41.53 & 9.14 & 28.96 & 44.59 & 27.43 \\
\multicolumn{1}{l}{} & CLIP-RSICD~\cite{pal2021Fine}\dag & ViT-B & RSICD & 151 & 14.09 & 30.10 & 43.64 & 11.16 & 33.25 & 48.91 & 30.19 \\
\multicolumn{1}{l}{} & RemoteCLIP~\cite{remoteclip} & ViT-B & RET-3 + DET-10 + SEG-4 & 151 & 17.02 & 37.97 & 51.51 & 13.71 & 37.11 & 54.25 & 35.26 \\
\multicolumn{1}{l}{} & RemoteCLIP~\cite{remoteclip} & ViT-L & RET-3 + DET-10 + SEG-4 & 428 & 18.39 & 37.42 & 51.05 & 14.73 & 39.93 & 56.58 & 36.35 \\
\multicolumn{1}{l}{} & GeoRSCLIP~\cite{zhang2024rs5m} & ViT-B & RS5M + RSICD + RSITMD & 151 & 15.59 & 41.19 & \textbf{57.99} & \textbf{21.13} & 41.72 & 55.63 & 38.87 \\

\multicolumn{1}{l}{} & \cellcolor{red!15}GeoMELT & \cellcolor{red!15}MWT &  \cellcolor{red!15} Multi-Task Data & \cellcolor{red!15}271 & \cellcolor{red!15}\textbf{22.87} & \cellcolor{red!15}\textbf{44.10} & \cellcolor{red!15}56.63 & \cellcolor{red!15}17.20 & \cellcolor{red!15}\textbf{43.59} & \cellcolor{red!15}\textbf{59.93} & \cellcolor{red!15}\textbf{40.72} \\
\midrule
\multirow{9}{*}{\rotatebox{90}{UCM}} & AMFMN~\cite{hoxha2020toward} & ResNet50 & UCM & 25* & 16.67 & 45.71 & 68.57 & 12.86 & 53.24 & 79.43 & 46.08 \\
 & KCR~\cite{mi2022knowledge} & ResNet18 & UCM & 11* & 11.90 & 48.57 & 71.43 & 17.24 & 56.95 & 81.14 & 47.87 \\
 & CLIP~\cite{radford2021Learninga}\dag & ViT-B & Zero-shot & 151 & 8.10 & 34.29 & 53.33 & 8.67 & 36.48 & 60.57 & 33.57 \\
 & CLIP~\cite{radford2021Learninga}\dag & ViT-L & Zero-shot & 428 & 11.91 & 43.33 & 65.24 & 10.76 & 46.00 & 73.33 & 41.76 \\
 & Rahhal et al.~\cite{rahhal2022multilanguage}\dag & ViT-B & UCM & 151 & 19.04 & 53.33 & 77.61 & 19.33 & 64.00 & 91.42 & 54.12 \\
 & CLIP-RSICD~\cite{pal2021Fine}\dag & ViT-B & RSICD & 151 & 15.71 & 50.00 & 82.38 & 13.81 & 57.05 & 91.24 & 51.70 \\
 & RemoteCLIP~\cite{remoteclip} & ViT-B & RET-3 + DET-10 + SEG-4 & 151 & \textbf{20.48} & \textbf{59.85} & 83.33 & 18.67 & 61.52 & 94.29 & 56.36 \\
 & RemoteCLIP~\cite{remoteclip} & ViT-L & RET-3 + DET-10 + SEG-4 & 428 & 19.05 & 54.29 & 80.95 & 17.71 & 62.19 & 93.90 & 54.68 \\

 \multicolumn{1}{l}{} & \cellcolor{red!15}GeoMELT & \cellcolor{red!15}MWT & \cellcolor{red!15} Multi-Task Data & \cellcolor{red!15}271 & \cellcolor{red!15}\textbf{20.48} & \cellcolor{red!15}57.14 & \cellcolor{red!15}\textbf{84.76} & \cellcolor{red!15}\textbf{18.86} & \cellcolor{red!15}\textbf{64.29} & \cellcolor{red!15}\textbf{94.95} & \cellcolor{red!15}\textbf{56.75} \\ \midrule
\multirow{11}{*}{\rotatebox{90}{RSITMD}} & AMFMN~\cite{hoxha2020toward} & ResNet50 & RSITMD & 25 & 10.63 & 24.78 & 41.81 & 11.51 & 34.69 & 54.87 & 29.72 \\
 & CMFM-Net~\cite{yu2022text} & ResNet18 & RSITMD & 11* & 10.84 & 28.76 & 40.04 & 10.00 & 32.83 & 47.21 & 28.28 \\
 & GaLR~\cite{yuan2022remote} & ResNet18 & RSITMD & 11* & 14.82 & 31.64 & 42.48 & 11.15 & 36.68 & 51.68 & 31.41 \\
 & CLIP~\cite{radford2021Learninga}\dag & ViT-B & Zero-shot & 151 & 9.74 & 22.57 & 34.51 & 8.72 & 27.88 & 42.88 & 24.38 \\
 & CLIP~\cite{radford2021Learninga}\dag & ViT-L & Zero-shot & 428 & 10.18 & 27.66 & 38.50 & 11.46 & 32.52 & 46.99 & 27.88 \\
 & Rahhal et al.~\cite{rahhal2022multilanguage} & ViT-B & RSITMD & 151 & 19.69 & 40.26 & 54.42 & 17.61 & 49.73 & 66.59 & 41.38 \\
 & CLIP-RSICD~\cite{pal2021Fine}\dag & ViT-B & RSICD & 151 & 25.00 & 46.46 & 61.73 & 19.29 & 51.02 & 68.45 & 45.32 \\
 & RemoteCLIP~\cite{remoteclip} & ViT-B & RET-3 + DET-10 + SEG-4 & 151 & 27.88 & 50.66 & 65.71 & 22.17 & 56.46 & 73.41 & 49.38 \\
 & RemoteCLIP~\cite{remoteclip} & ViT-L & RET-3 + DET-10 + SEG-4 & 428 & \textbf{28.76} & 52.43 & 63.94 & 23.76 & 59.51 & 74.73 & 50.52 \\
 & GeoRSCLIP~\cite{zhang2024rs5m} & ViT-B & RS5M + RSICD + RSITMD & 151 & 25.04 & \textbf{57.88} & \textbf{74.38} & \textbf{32.30} & 53.32 & 67.92 & \textbf{51.81} \\
 \multicolumn{1}{l}{} & \cellcolor{red!15}GeoMELT & \cellcolor{red!15}MWT & \cellcolor{red!15} Multi-Task Data & \cellcolor{red!15}271 & \cellcolor{red!15}25.66 & \cellcolor{red!15}53.98 & \cellcolor{red!15}67.70 & \cellcolor{red!15}25.80 & \cellcolor{red!15}\textbf{60.35} & \cellcolor{red!15}\textbf{76.24} & \cellcolor{red!15}51.62 \\ \bottomrule
\end{tabular}
}
\end{table*}

Regarding text-image retrieval, GeoMELT is used as a dual-encoder, with each modality input being passed separately to the model, obtaining $F_\theta(\mathbf{x}_i)=\mathbf{v}_i \in \mathbb{R}^D$ and $G_\theta(\mathbf{y}_i)=\mathbf{u}_i \in \mathbb{R}^D$, with $D$ corresponding to the hidden embedding dimensionality of the encoder. These vector representations are aligned through the use of the InfoNCE loss, as typically used in CLIP, so that given a batch of $N$ image-text pairs, the similarity between correct pairs is maximized while minimizing the similarity between pairs of other in-batch elements. During inference, retrieved elements can be selected by their cosine similarity with the query.

Formally, the InfoNCE loss $\mathcal{L}_{\mathrm{InfoNCE}}$ is minimized as the sum of both the text-to-image $\mathcal{L}_{\mathrm{t2i}}$ and image-to-text directions $\mathcal{L}_{\mathrm{i2t}}$, as described in the following equations:

\begin{equation} \label{infonceloss_t2i}
    \mathcal{L}_{\mathrm{t2i}} = - \frac{1}{N} \sum^{N}_{i=1} \left( \log \frac{\exp( \cos(\mathbf{u}_i, \mathbf{v}_i)/\tau)}{\sum^{N}_{j=1}\exp(\cos(\mathbf{u}_i,\mathbf{v}_i)/\tau)} \right),
\end{equation}

\begin{equation} \label{infonceloss_i2t}
    \mathcal{L}_{\mathrm{i2t}} = - \frac{1}{N} \sum^{N}_{i=1} \left( \log \frac{\exp( \cos(\mathbf{v}_i,\mathbf{u}_i)/\tau)}{\sum^{N}_{j=1}\exp(\cos(\mathbf{v}_i,\mathbf{u}_i)/\tau)} \right),
\end{equation}
\begin{equation}
    \mathcal{L}_{\mathrm{InfoNCE}} = \frac{\mathcal{L}_{\mathrm{t2i}} + \mathcal{L}_{\mathrm{i2t}}}{2}.
\end{equation}

Notice that both text generation and cross-modal retrieval are optimized jointly, and our complete data mixture includes datasets for tasks such as VQA and VG, which either lack image–caption pairs or provide captions of limited quality. In these cases, we exclude such instances from the training batches used for the retrieval objective.

\subsection{Multi-Task Learning }

We now detail the considered strategies for jointly optimizing the text generation and cross-modal retrieval objectives. 

\subsubsection{Multiple Loss Balancing}

We combine the multiple losses using Dynamic Weight Average (DWA)~\cite{liu2019end}. DWA is a lightweight method that adaptively estimates task weights based on the rate of change of the training losses. This will result in higher weights for tasks whose losses decrease slowly, while tasks that decrease faster are assigned lower weights. For each task $i \in \{\mathrm{MLM}, \mathrm{InfoNCE} \}$ and epoch $k$, the ratio of the loss change is calculated as $\omega_i^{(k-1)} = \frac{\mathcal{L}_i^{(k-1)}}{\mathcal{L}_i^{(k-2)}}$.

Given these change ratios, with $m=2$ tasks and a temperature parameter $\gamma$, the weight associated for task $i$ at epoch $k$ is calculated as:

\begin{equation}
    \lambda_i^{(k)} = 
    \frac{m \, \exp\!\big(\omega_i^{(k-1)}/\gamma\big)}
         {\sum_{j=1}^{m} \exp\!\big(\omega_j^{(k-1)}/\gamma\big)}.
\end{equation}

The final training loss can then be calculated as $\mathcal{L} = 
\lambda_{\mathrm{MLM}} \, \mathcal{L}_{\mathrm{MLM}} +
\lambda_{\mathrm{InfoNCE}} \, \mathcal{L}_{\mathrm{InfoNCE}}$, with both $\lambda_i$ learnable parameters evolving during training.



\subsubsection{Two-Stage Training}

In our early experiments, we noticed that training the model solely on the visual grounding task resulted in strong performance for that task. However, when applying the multi-task training procedure, the performance of this task remained low. To address this issue, we introduced a warm-up phase in which the model was first trained exclusively on the visual grounding task before applying multi-task learning, which resulted in improved final performance across the multiple tasks. Additionally, we also observed that our model benefited from initialization with a BEiT3 checkpoint finetuned for the retrieval task over generalist images. To benefit from both initializations, we adopted a simple technique from WISE-FT~\cite{wortsman2022robust}, where each parameter is obtained as a weighted average of the two checkpoints. This approach is simple and free, as the BEiT3 retrieval checkpoint is already publicly available. These strategies are empirically justified in Section~\ref{section:ckpt-init}.

\begin{figure*}[t]
  \centering
   \includegraphics[width=\linewidth]{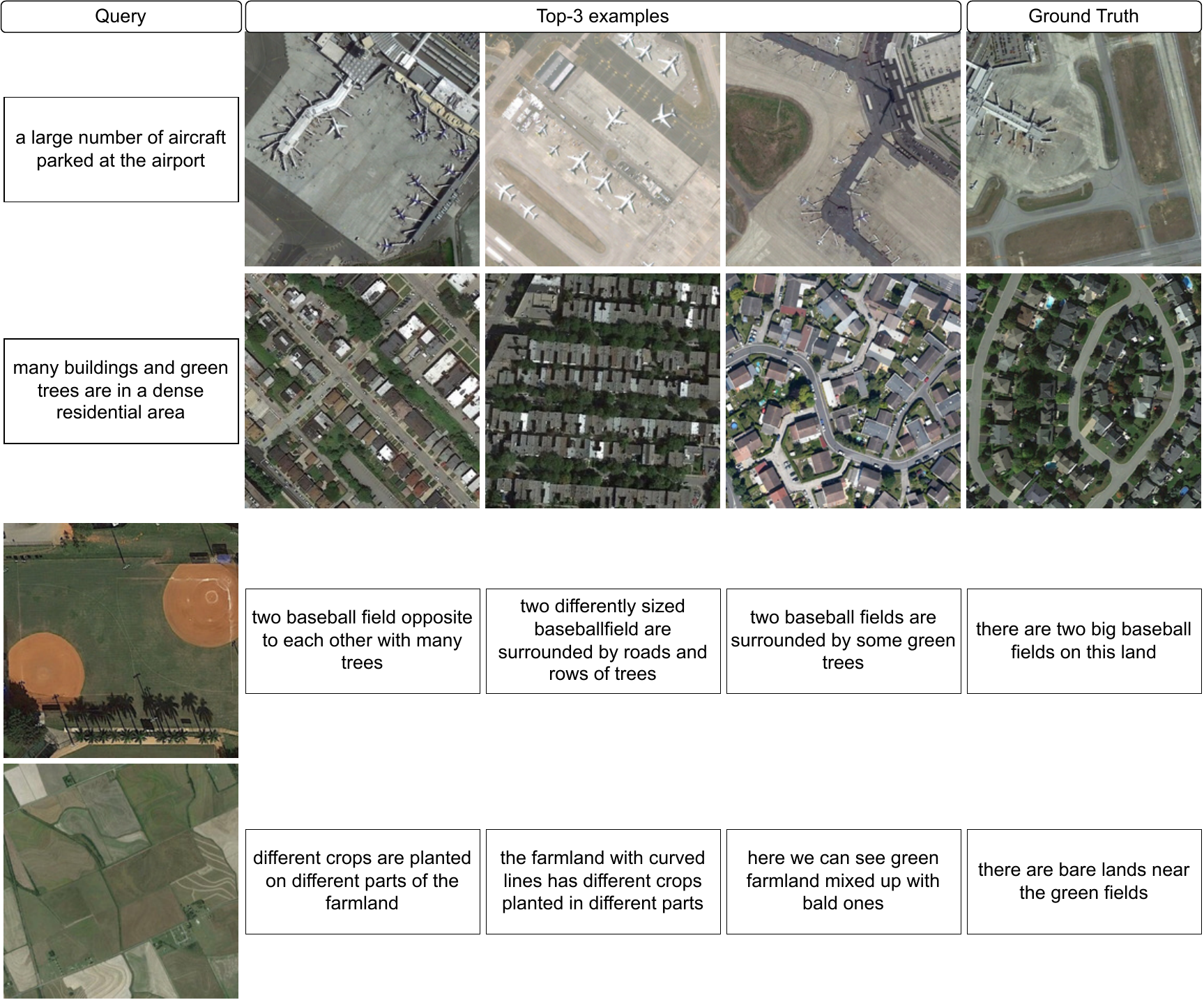}

   \caption{Examples of image–text retrieval. In some cases, GeoMELT produces results that are more semantically aligned than the ground truth. In the first row, the ground-truth image does not contain the presence of multiple airplanes, which GeoMELT correctly identifies. In the last row, GeoMELT obtains a more detailed and contextually rich description of the image.
   }
   \label{fig:retrieval}
\end{figure*}

\section{Experimental Setup}\label{experimental}

This section describes the experimental setup, including the datasets used for training, implementation details, and metrics used for the evaluation of the proposed approach.

\subsection{Datasets}

The following is a description of each dataset that was included in the data mixture for training GeoMELT:

\begin{itemize}
    \item NWPU-Captions~\cite{cheng2022NWPUCaptions} is a large dataset with manually curated captions, consisting of 31,500 images with 5 sentences each, forming a total of 157,500 pairs. Images are equally divided across 45 categories, covering different types of land use and land coverage. The spatial resolution of the images ranges from 0.2 to 30 meters.
    \item RSICD~\cite{lu2017exploring} was collected from different image sources covering 30 different classes, with 10,921 total images paired with manually annotated short captions, of which many are nonetheless duplicated.
    \item UCM-Captions and Sydney-Captions are smaller datasets that were repurposed from scene classification by manual annotation of the captions~\cite{qu2016deep}. Despite their widespread use, the captions from these datasets are of lower quality, with a high number of text duplicates as well.
    \item RSITMD~\cite{yuan2022exploring} is a dataset of 23,715 images proposed to address the limitations of previous retrieval datasets, featuring fine-grained information and a greater difference between sentences.
\end{itemize}

We also incorporate datasets containing longer and more detailed text descriptions to enhance the capability of GeoMELT to generate richer captions and improve retrieval performance for longer textual queries.

\begin{itemize}
    \item RSICap~\cite{hu2025rsgpt} is a high-quality dataset of 2,585 human-annotated image-text pairs, featuring examples of detailed descriptions with an average caption length of 60 words. 
    \item VRSBench~\cite{li2024vrsbench} contains 29,614 image–text pairs with longer and more fine-grained captions, constructed through a semi-automatic pipeline and humanly verified. Each caption has an average of 4 sentences and 52 words. 
\end{itemize}

To prevent data leakage when combining multiple text–image retrieval datasets, we adopt the filtered versions of RSICD, UCM, and RSITMD provided by Liu et al.~\cite{remoteclip}. Additionally, in order to extend the capabilities of GeoMELT, we incorporate datasets from other tasks beyond captioning:

\begin{itemize}
    \item Visual Question Answering: RSVQA-HR~\cite{lobry2020RSVQA} is a dataset containing $\sim$1M image, question, and answer triplets. For a balanced proportion with the other datasets, we follow previous approaches~\cite{pang2025vhm, muhtar2024lhrs} and randomly sample 20,000 instances from the training split. Additionally, we include 10,000 randomly selected examples from the RSVQA-LR dataset, which uses Sentinel 2 images instead of very high resolution ones, therefore theoretically widening the scale robustness of GeoMELT.
    \item Visual Grounding: DIOR-RSVG~\cite{zhan2023rsvg} is a grounding dataset where images are grouped with object descriptions and their respective coordinates. 26,991 instances are used for training, with 7,500 instances for testing.
\end{itemize}

We have aggregated the information about the training data in Table~\ref{tab:datasets}. As shown there, GeoMELT is trained with only a fraction of the total volume of data considered by other relevant approaches, including models based on CLIP (RemoteCLIP, GeoRSCLIP) and LVLMs (SkyEyeGPT and LHRS-Bot), thus significantly lowering the overall training cost.


\begin{table*}[t!]
\centering
\caption{Results over multiple image captioning benchmarks, with * indicating vision-only parameter counts.}
\resizebox{\textwidth}{!}{
\begin{tabular}{lccccccccccccc}
\toprule
\multirow{2}{*}{Method} & \multirow{2}{*}{\shortstack{\\Total\\Params}} & \multicolumn{3}{c}{RSICD} & \multicolumn{3}{c}{UCM-Captions} & \multicolumn{3}{c}{Sydney-Captions} & \multicolumn{3}{c}{NWPU-Captions} \\ \cmidrule{3-5} \cmidrule{6-8} \cmidrule{9-11} \cmidrule{12-14}
 & & ~BLEU1~ & ~BLEU4~ & ~CIDEr~ & ~BLEU1~ & ~BLEU4~ & ~CIDEr~ & ~BLEU1~ & ~BLEU4~ & ~CIDEr~ & ~BLEU1~ & ~BLEU4~ & ~CIDEr~ \\ \midrule
MLCA-NET~\cite{cheng2022NWPUCaptions} & 138M* & 0.757 & 0.461 & 2.356 & 0.826 & 0.668 & 3.240 & 0.831 & 0.580 & 2.324 & 0.745 & 0.478 & 1.164 \\
HCNet~\cite{yang2024hcnet}     & 25M* & - & - & - & 0.883 & 0.745 & 3.518 & 0.769 & 0.610 & 2.471 & 0.896 & 0.717 & 2.093 \\
MC-Net~\cite{huang2023mc}             & 138M* & 0.728 & 0.433 & 2.454 & 0.845 & 0.679 & 3.355 & 0.834 & 0.607 & 2.564 & 0.741 & 0.478 & 1.159 \\
BITA~\cite{yang2024bootstrapping}     & 3B & 0.774 & 0.504 & \textbf{3.054} & 0.889 & 0.719 & 3.845 & - & - & - & 0.885 & 0.676 & 1.970   \\
Aware-Trans.~\cite{cao2023aware}      & 28M* & -  & - & - & 0.901 & \textbf{0.781} & 3.779 & 0.854 & 0.651 & 2.832 & \textbf{0.915} & \textbf{0.750} & 2.147 \\
Def. Trans.~\cite{du2023deformable}   & 90M* & 0.758 & 0.492 & 2.581 & 0.823 & 0.679 & 3.463 & 0.837 & 0.666 & \textbf{3.037} & 0.752 & 0.483 & 1.207 \\ \midrule
RSGPT~\cite{hu2025rsgpt} & 13B & 0.703 & 0.368 & 1.029 & 0.861 & 0.657 & 3.332 & 0.823 & 0.622 & 2.731 & - & - & -  \\
RS-CapRet~\cite{silva2024large} & 7B & 0.741 & 0.455 & 2.605 & 0.833 & 0.645 & 3.429 & 0.782 & 0.545 & 2.390 & 0.871 & 0.656 & 1.929  \\
SkyEyeGPT~\cite{zhan2025skyeyegpt} & 7B &  \textbf{0.867} & \textbf{0.600} & 0.837 & \textbf{0.907} & 0.784 & 2.368 & \textbf{0.919} & \textbf{0.774} & 1.811 & - & -  & - \\ 
EarthGPT~\cite{zhang2024earthgpt} & 7B & - & - & - & - & - & - & - & - & - & 0.871 & 0.655  & 1.926 \\ \midrule
\cellcolor{red!15}GeoMELT & \cellcolor{red!15}271M & \cellcolor{red!15}0.610 & \cellcolor{red!15}0.365 & \cellcolor{red!15}2.652 & \cellcolor{red!15}0.898 & \cellcolor{red!15}0.772 & \cellcolor{red!15}\textbf{3.917} & \cellcolor{red!15}0.791 & \cellcolor{red!15}0.575 & \cellcolor{red!15}2.643 & \cellcolor{red!15}0.907 & \cellcolor{red!15}0.731 & \cellcolor{red!15}\textbf{2.155} \\ \bottomrule

\end{tabular}
}
\label{table:captioning}
\end{table*}

\subsection{Implementation Details}

The parameters of GeoMELT are initialized from a publicly available BEiT3 checkpoint~\cite{wang2023image}, which was trained with masked data modeling (masking both text tokens and image patches). In total, the model has 271 million parameters. Input images are resized to $224\times224$ pixels, with patches of size 16. In the first stage, we finetune GeoMELT on the visual grounding task over 30 epochs, with a batch size of 64, a learning rate of $4\times 10^{-5}$, and 1 warm-up epoch. In the second stage, we aggregate the multiple remote sensing datasets and perform multi-task training on the text generation and image-text retrieval tasks, for 15 epochs with a batch size of 128 and a learning rate of $2\times10^{-4}$. AdamW was used as the optimizer for both stages. The temperature parameter $\gamma$ was set to 2.0, and the $\alpha$ parameter for WISE-FT was set to 0.5.

\subsection{Metrics}


\begin{table}[t!]
\centering
\caption{Comparison of different methods on the long captioning task, over the VRSBench dataset.}
\begin{tabularx}{\columnwidth}{l Y Y Y Y}
\toprule
Method & Total Params & BLEU-1 & BLEU-4 & CIDEr \\
\midrule
\multicolumn{5}{l}{\cellcolor{gray!15}\textit{Zero-shot}} \\
GeoChat~\cite{kuckreja2024geochat} & 7B & 13.9 & 1.4 & 0.4 \\
GPT-4V~\cite{achiam2023gpt}        & UNK & 37.2 & 8.6 & 19.1 \\
\midrule
\multicolumn{5}{l}{\cellcolor{gray!15}\textit{Finetuned}} \\
MiniGPT-v2~\cite{chen2023minigpt}     & 7B & 36.8 & 8.7 & 21.4 \\
LLaVA-1.5~\cite{liu2024improved}     & 7B & \textbf{48.1} & \textbf{14.7} & \textbf{33.9} \\
GeoChat~\cite{kuckreja2024geochat} & 7B & 46.7 & 13.8 & 28.2 \\
Mini-Gemini~\cite{li2024mini} & 7B & 47.6 & 14.3 & 33.5 \\
\cellcolor{red!15}GeoMELT    & \cellcolor{red!15}241M & \cellcolor{red!15}45.8 & \cellcolor{red!15}13.8 & \cellcolor{red!15}29.8 \\
\bottomrule
\end{tabularx}
\label{tab:longcap}
\end{table}

We use standard image captioning metrics for the evaluation of our model, mainly BLEU and CIDEr, which measure the overlap of n-grams of the generated and reference captions. For retrieval tasks, we use R@k and mR, where R@k means the fraction of queries for which the relevant item is ranked among the top-k retrieved elements, and mR is the average of all R@k for both text-image and text-image retrieval. As for visual grounding, we use mean Intersection over Union (mIoU) with a threshold of 0.5, measuring the ratio of the overlapping area between the predicted bounding box coordinates and the ground truth. For VQA, we present the overall accuracy along with the average accuracy by answer type.

\section{Experimental Results}\label{results}

We have conducted extended evaluations of GeoMELT in common benchmarks spanning multiple remote sensing tasks.

\subsection{Main Experimental Results}

\subsubsection{Text-Image Retrieval} We start with text–image retrieval results, with comparisons against several baselines summarized in Table~\ref{table:clip-objective}. These baselines can be divided into two groups. The first group consists of methods with specialized architectures and a relatively lower number of parameters, trained exclusively on the corresponding training split. The second group consists of larger models, leveraging vision and language backbones based mainly on CLIP~\cite{radford2021Learninga} and finetuned for image-text retrieval in remote sensing.

Firstly, it can be observed that CLIP models are strong baselines for cross-modal retrieval in remote sensing, outperforming even specialized methods in a zero-shot setting. These improvements come from their larger model capacity and larger quantity of training data. Multiple studies have demonstrated the benefits of leveraging pretrained representations from these CLIP models~\cite{remoteclip,zhang2024rs5m}. GeoMELT achieves strong performance compared with these approaches, consistently improving over RemoteCLIP across multiple datasets, including its larger variant trained on substantially more data (e.g., +4.37 mR in RSICD and +1.1 mR in RSITMD). GeoMELT also achieves superior results on RSICD (+ 1.85 mR) compared to GeoRSCLIP~\cite{zhang2024rs5m}, while being competitive in RSITMD (- 0.19 mR). These findings confirm the effectiveness of our architecture for cross-modal retrieval, despite being trained with a modest amount of data (see Table~\ref{tab:datasets} for a comparison between training data amounts).

Some examples of image-text retrieval are presented in Fig.~\ref{fig:retrieval}. In several instances, GeoMELT retrieves samples that demonstrate stronger semantic alignment than the corresponding ground-truth annotations. For example, in the first row, the query \textit{"a large number of aircraft parked at the airport"} is not correctly paired with an image containing multiple airplanes in the ground-truth set. However, GeoMELT understands this detail from the query and obtains multiple relevant images. In the second row, several reasonable examples are retrieved for a query describing a residential area. In the case of text retrieval given an input image, GeoMELT retrieves captions that offer richer and more detailed descriptions than the original annotations. These results indicate that, despite the presence of noise and vagueness in the datasets, GeoMELT effectively aligns semantically relevant images and captions.

\subsubsection{Image Captioning} In Table~\ref{table:captioning} we report results over the image captioning task, comparing GeoMELT against a variety of approaches that include earlier encoder-decoder models, transformer-based methods, and LVLMs. GeoMELT achieves strong performance across datasets and metrics, outperforming other LVLMs and remaining competitive or surpassing other approaches trained specifically for the task. In particular, the CIDEr metric of GeoMELT is 2.652 mR in RSICD (only behind 3.054 of BITA~\cite{yang2024bootstrapping}), 3.917 in UCM  (SOTA), 2.643 in Sydney, and 2.155 in NWPU-Captions (SOTA). These results demonstrate that the compact architecture of GeoMELT can effectively describe images from different sources, often comparing favourably against other LVLMs that have many more parameters (over 7 billion).

For the detailed image captioning task, GeoMELT achieves a performance comparable to that of most finetuned baselines. In particular, compared to another remote sensing model, namely the GeoChat model, GeoMELT has a higher CIDEr score (+1.6), with a slight decrease in BLEU-1 (-0.9) and an identical BLEU-4 score.

\begin{table}[t]
\centering
\caption{Comparison of different methods on RSVQA-HR.}
\begin{tabularx}{\columnwidth}{l Y Y Y Y}
\toprule
Method & Total Params & HR-presence & HR-compare & Avg. \\ \midrule
LLaVA-1.5~\cite{liu2024improved} & 7B & 69.83 & 67.29 & 68.56 \\
MiniGPTv2~\cite{chen2023minigpt} & 7B & 40.79 & 50.91 & 46.46 \\
Qwen-VL-Chat~\cite{Qwen-VL} & 9.6B & 66.44 & 60.41 & 63.06 \\
GeoChat~\cite{kuckreja2024geochat} & 7B & 58.45 & 83.19 & 72.30 \\
EarthGPT~\cite{zhang2024earthgpt} & 7B & 62.77 & 79.53 & 71.15 \\
VHM~\cite{pang2025vhm}& 7B & 64.00 & \textbf{83.50} & 73.75 \\
EarthDial~\cite{soni2025earthdial} & 4B & 58.89 & 83.11 & 72.45 \\
GeoVLM-R1~\cite{fiaz2025geovlmr1} & 3B & 66.38 & 82.26 & \textbf{75.27} \\
\cellcolor{red!15}GeoMELT & \cellcolor{red!15}271M & \cellcolor{red!15}\textbf{70.16} & \cellcolor{red!15}66.61 & \cellcolor{red!15}68.19 \\ \bottomrule
\end{tabularx}
\label{tab:rsvqa-hr}
\end{table}

\begin{table}[t!]
\centering
\caption{Comparison of different methods on RSVQA-LR.}
\begin{tabularx}{\columnwidth}{l Y Y Y Y Y}
\toprule
Method & Total Params & LR-rural & LR-presence & LR-compare & Avg. \\
\midrule
LLaVA-1.5~\cite{liu2024improved} & 7B & 59.00 & 55.46 & 68.20 & 62.77  \\
MiniGPTv2~\cite{chen2023minigpt} & 7B & 39.00 & 55.16 & 55.22 & 54.96  \\
Qwen-VL-Chat~\cite{Qwen-VL} & 9.6B & 61.00  &38.57 & 67.59 & 55.35 \\
RSGPT~\cite{hu2025rsgpt} & 13B & 94.00 & 91.17 & 91.70 & 92.29 \\
GeoChat~\cite{kuckreja2024geochat} & 7B & 94.00 & 91.09 & 90.33 & 91.81 \\
SkyEyeGPT~\cite{zhan2025skyeyegpt} & 7B & 75.00 & 88.93 & 88.63 & 84.19 \\
LHRS-Bot~\cite{muhtar2024lhrs} & 7B & 89.07 & 88.51 & 90.00 & 89.19 \\
VHM~\cite{pang2025vhm} & 7B & 88.00 & 90.11 & 89.89 & 89.33 \\
EarthDial~\cite{soni2025earthdial} & 4B & 92.75 & \textbf{94.00} & 92.58 & 92.70 \\
GeoVLM-R1~\cite{fiaz2025geovlmr1} & 3B & \textbf{96.00} & 91.81 & \textbf{93.20} & \textbf{92.66} \\
\cellcolor{red!15}GeoMELT & \cellcolor{red!15}271M & \cellcolor{red!15}32.00 & \cellcolor{red!15}78.34 & \cellcolor{red!15}66.44 & \cellcolor{red!15}70.94 \\
\bottomrule
\end{tabularx}
\label{tab:rsvqa-lr}
\end{table}

\begin{figure}[t]
  \centering
   \includegraphics[width=\linewidth]{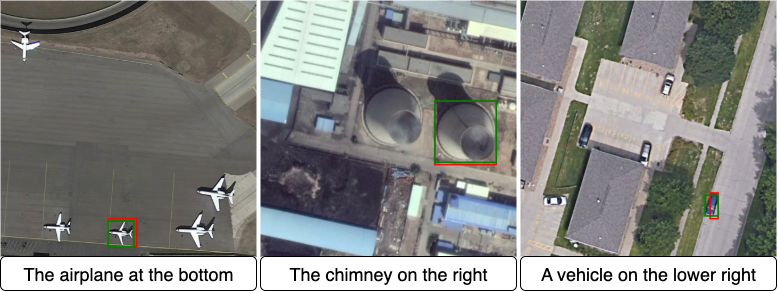}

   \caption{Examples of visual grounding outputs given textual queries. GeoMELT can localize the objects of interest and successfully differentiate them based on localization.
   }
   \label{fig:vg}
\end{figure}

\begin{table}[t!]
\centering
\caption{Comparison of different methods on the DIOR-RSVG benchmark for visual grounding. 
}
\begin{tabularx}{\columnwidth}{l Y Y Y}
\toprule
Method & Total Params &  Input Size & mIoU \\
\midrule
CogVLM~\cite{wang2024cogvlm} & 17B & $490\times490$ & 44.58 \\
Qwen-VL-Chat~\cite{Qwen-VL} & 9.6B & $448\times448$ & 31.86 \\
\midrule
VHM~\cite{pang2025vhm} & 7B & $336\times336$ & 56.17 \\
EarthGPT~\cite{pang2025vhm} & 7B & $518\times518$ & \textbf{69.34} \\ 
\cellcolor{red!15}GeoMELT & \cellcolor{red!15}271M & \cellcolor{red!15}$224\times224$ & \cellcolor{red!15}65.95 \\
\bottomrule
\end{tabularx}
\label{table:dior-rsvg}
\end{table}

\subsubsection{Visual Grounding} Visual grounding results can be seen in Table~\ref{table:dior-rsvg}. GeoMELT achieves a mean IoU of 65.95 and outperforms both general-purpose LVLMs and remote sensing models, such as VHM~\cite{pang2025vhm}, while operating at a significantly lower image resolution (i.e., $224\times224$) than the others. EarthGPT achieves higher performance, but at the cost of using a larger input resolution of $518\times518$ pixels and by incorporating multi-layer visual features. The efficiency of the compact design of GeoMELT is thus evidenced in this task, not only by its much lower number of parameters, but also by its lower input image resolution, which is particularly advantageous given the quadratic complexity of the self-attention mechanism. Fig.~\ref{fig:vg} presents multiple examples of bounding box predictions given query object descriptions. The results demonstrate that the model can accurately localize the referenced objects and distinguish between multiple instances of the same category when provided with contextual cues.

\subsubsection{Visual Question Answering} As for visual question answering, the performance of GeoMELT compared to other LVLMs is shown in Table~\ref{tab:rsvqa-hr}. We follow previous work in omitting area and count questions~\cite{kuckreja2024geochat,pang2025vhm}. It can be seen that GeoMELT is quite competitive regarding the \textit{presence} category, but falls short in the \textit{compare} category, which requires more reasoning abilities. The advantage of LVLMs is more evident in this task, which we hypothesize comes from the LLM component's ability to better model the relationships between the objects referenced in the questions. Still, GeoMELT performs competitively, with an average score of 68.19. The performance of GeoMELT declines more in the RSVQA-LR dataset, with an average of 70.94, possibly due to the nature of the images being from a much lower resolution than the rest of the data mixture used for training, therefore showing that full scale invariance is not achievable by our small multimodal approach. We believe that adding more low-resolution data would be necessary in training to improve results.

\begin{table*}[t!]
\centering
\caption{Zero-shot image classification in 12 remote sensing datasets.}
\resizebox{\textwidth}{!}{
\begin{tabular}{lcccccccccccccc}
\toprule
\multicolumn{1}{c}{Method} & Backbone & \rotatebox{45}{RSI-CB128} &
\rotatebox{45}{RSI-CB256} &
\rotatebox{45}{WHU-earth} &
\rotatebox{45}{EuroSAT} &
\rotatebox{45}{MLRSNet} &
\rotatebox{45}{PatternNet} &
\rotatebox{45}{RESISC45} &
\rotatebox{45}{AID} &
\rotatebox{45}{RSSCN7} &
\rotatebox{45}{OPTIMAL-31} &
\rotatebox{45}{RSC11} &
\rotatebox{45}{WHU-RS19} &
\rotatebox{45}{Average} \\ \midrule
\multicolumn{15}{l}{\cellcolor{gray!15}\textit{CLIP-based Methods}} \\
CLIP~\cite{radford2021Learninga} & ViT-B & 20.09 & 29.40 & 40.62 & 41.83 & 41.12 & 49.65 & 50.65 & 51.40 & 65.00 & 66.13 & 46.40 & 73.79 & 48.01 \\
CLIP~\cite{radford2021Learninga} & ViT-L & 32.69 & 43.65 & 56.46 & 39.19 & 59.09 & \textbf{70.15} & 64.16 & 62.70 & 67.68 & 73.92 & 64.26 & 87.86 & 60.15 \\
RemoteCLIP~\cite{remoteclip} & ViT-B & 25.88 & 42.56 & 63.12 & 33.04 & 57.91 & 58.01 & 66.87 & 86.60 & 73.75 & 79.03 & 57.16 & 94.17 & 61.51 \\
RemoteCLIP~\cite{remoteclip} & ViT-L & \textbf{37.23} & \textbf{51.55} & 68.33 & 46.04 & 61.67 & 63.12 & 73.68 & 76.95 & 71.07 & 80.65 & 68.43 & 92.72 & 65.95 \\ 
\midrule
\multicolumn{15}{l}{\cellcolor{gray!15}\textit{Large Vision-Language Models}} \\
GeoChat~\cite{kuckreja2024geochat} & LLaVA-1.5-7B &  &  &  &  &  &  &  & 72.03 &  &  &  & 80.09  &  
\\ 
EarthDial~\cite{soni2025earthdial} & Phi-3 &  &  &  &  &  &  &  & \textbf{88.76} &  &  &  & 96.21 &  
\\ 
GeoVLM-R1~\cite{fiaz2025geovlmr1} & Qwen2.5VL-3B &  &  &  &  &  &  &  & 88.46 &  &  &  & 97.91 &  
\\ \midrule
\cellcolor{red!15}GeoMELT & \cellcolor{red!15} BEiT-3 & \cellcolor{red!15}36.13 & \cellcolor{red!15}46.45 & \cellcolor{red!15}\textbf{78.33} & \cellcolor{red!15}\textbf{49.19} & \cellcolor{red!15}\textbf{73.21} & \cellcolor{red!15}64.01 & \cellcolor{red!15}\textbf{96.81} & \cellcolor{red!15}87.40 & \cellcolor{red!15}\textbf{78.04} & \cellcolor{red!15}\textbf{98.66} & \cellcolor{red!15}\textbf{78.58} & \cellcolor{red!15}\textbf{98.54} & \cellcolor{red!15}\textbf{73.78} \\
\bottomrule
\end{tabular}
}
\label{table:classification}
\end{table*}

\begin{table*}[t]
\centering
\caption{Results from ablating different checkpoint initializations.}
\begin{tabular}{lccccccccc}
\toprule
\multirow{2}{*}{\shortstack{\\Checkpoint\\Initialization}} & \multirow{2}{*}{\shortstack{\\Finetuning\\Strategy}} & \multicolumn{3}{c}{Image-Text Retrieval (mR)} & \multicolumn{4}{c}{Image Captioning (CIDEr)} & V. Grounding \\ \cmidrule{3-10} 
                                   & \multicolumn{1}{l}{}                                     & RSICD          & UCM         & RSITMD         & RSICD      & UCM     & Sydney     & NWPU     & DIOR-RSVG    \\ \midrule
Retrieval BEiT3                & Zero-shot                                                & 21.48	&41.81	& 31.91 & - & - & - & - & - \\
Captioning BEiT3               & Zero-shot & - & - & - & 0.114	 & 0.148 & 0.067	& 0.075 & - \\ \midrule
Captioning BEiT3               & VG only                                         & -              & -           & -              & -          & -       & -          & -        & \cellcolor{green!60!black}66.17\\
Retrieval BEiT3                & GeoMELT & \textbf{41.33} & 57.57 & \textbf{54.54} & \textbf{2.793} &	3.830	& \textbf{2.579}	&2.095& \cellcolor{red!50}15.97 \\
Retrieval BEiT3                & GeoMELT (w/ +10 epochs) & 40.76 & \textbf{57.98} & 54.09 &2.751	& \textbf{3.965}	&2.505	& \textbf{2.148}& \cellcolor{red!50}15.12 \\
Visual Grounding (VG)                             & GeoMELT & 35.22 & 56.37 & 47.77 &2.676	&     3.552	    &   2.502       &	2.074 & \cellcolor{green!60!black}\textbf{66.52} \\
WISE-FT (VG + Retrieval)           & GeoMELT & 39.32 & 57.48 & 51.95 & 2.678& 	3.885	&2.523	&2.122 & \cellcolor{green!60!black}\textbf{66.52} \\ \bottomrule
\end{tabular}
\label{tab:ablations}
\end{table*}

\subsubsection{Zero-shot Image Classification} Finally, we discuss the results of GeoMELT over multiple zero-shot image classification tasks, which are compiled in Table~\ref{table:classification}. For each class $c$ in each dataset, a prompt in the form of "\texttt{a satellite photo of} $c$" is constructed. When using the model as a dual encoder, the corresponding text embeddings are obtained, and an input image is then assigned to the class whose text embeddings have the highest cosine similarity with the image embedding. Our evaluation includes comparisons with existing approaches in the remote sensing domain that leverage CLIP or LVLM-based architectures. The results indicate that our model performs robustly across multiple datasets, with an average score of 73.78, significantly improving the score from RemoteCLIP-Large (65.95).

\subsection{Checkpoint Initialization Experiments}\label{section:ckpt-init}

To evaluate the effectiveness of our initialization strategy, we performed a series of experiments summarized in Table~\ref{tab:ablations}. We adopted a simplified data mixture comprising the short image captioning, RSVQA-HR, DIOR-RSVG, and RSICap datasets, reporting results for the retrieval, captioning, and visual grounding tasks. As a baseline, we measure the results of the original checkpoints. The results of the retrieval model are quite reasonable (e.g., mR in UCM of 41.81), while, as expected, the performance for captioning is very low (e.g., 0.075 CIDEr in NWPU). We started the training by using a default initialization from the BEIT3 checkpoint finetuned on the retrieval task. While this approach achieves strong performance on retrieval and captioning (e.g., 54.54 mR in RSITMD and 2.793 in RSICD), the model performs poorly on the visual grounding task, with a mean IoU of 15.97. This limitation persists even when the model is trained with more epochs, with similar results in the retrieval and captioning tasks (e.g., 54.09 mR in RSITMD and 2.751 in RSICD). 

In a second experiment, we trained our model in the visual grounding task \textit{exclusively}, which achieves a mean IoU of 66.17, and then used this as an initialization checkpoint. This approach can achieve the desired performance for the visual grounding task (mIoU of 66.52) while also demonstrating high performance in the retrieval and captioning tasks, albeit slightly lower (e.g., 47.77 mR in RSITMD and 2.676 in RSICD). Finally, we showcase the benefits of applying a simple linear model merging (WISE-FT~\cite{wortsman2022robust}) for the initialization, which allows GeoMELT to regain performance results across retrieval and image captioning tasks (e.g., 51.95 mR in RSITMD and 2.678 in RSICD).

\section{Multitask Examples}

\begin{figure*}[t!]
  \centering
   \includegraphics[width=0.7\linewidth]{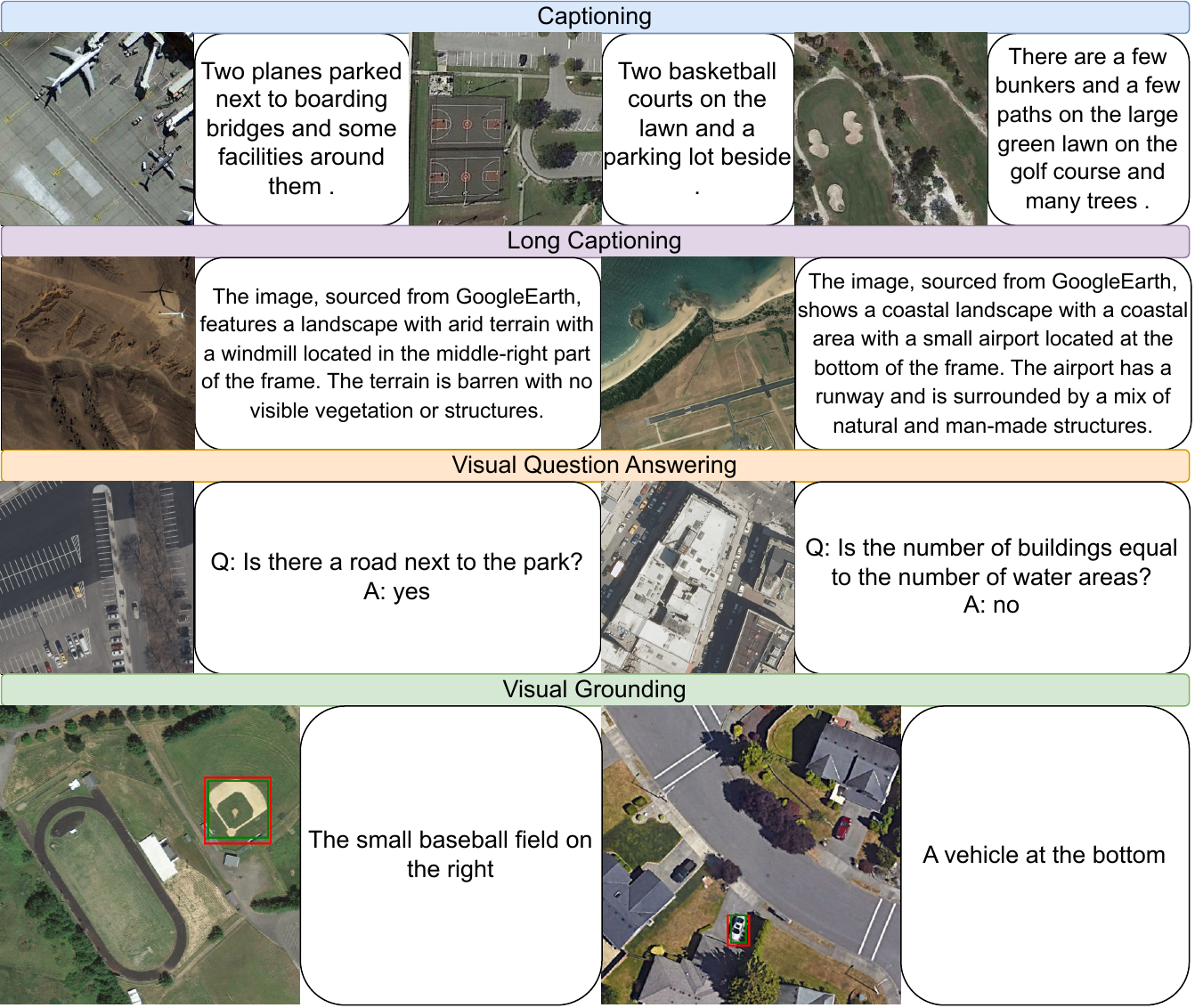} 

   \caption{Examples of GeoMELT outputs across different tasks, including captioning, long captioning, visual question answering, and visual grounding. GeoMELT produces concise descriptions, generates detailed long captions, accurately answers questions from RSVQA datasets, and successfully identifies queried objects.
   }
   \label{fig:qualitative}
\end{figure*}



We present illustrative examples for the model outputs in Fig.~\ref{fig:qualitative}, showing that GeoMELT is capable of addressing multiple tasks successfully. The first row presents examples of short descriptions, similar to those in typical remote sensing image captioning datasets. In contrast, the second row shows two cases where the model generates descriptions that provide greater detail regarding the objects mentioned and their localization within the image. GeoMELT goes into detail regarding the surface terrain and the localization of objects of interest. Some VQA examples from the RSVQA-HR test set are also shown, demonstrating the ability of GeoMELT to answer questions based on the image contents. 


Overall, the results show that although GeoMELT is a lightweight model, it can still provide reasonable responses across different types of requests, challenging state-of-the-art models that are computationally more demanding.

\section{Conclusions and Future Work}\label{conclusion}

This paper introduced a lightweight model named GeoMELT, based on a multimodal encoder, that can successfully address multiple vision and language tasks with remote sensing data. We show empirically that the architecture of GeoMELT is well-suited for both image-text retrieval and text generation tasks, which are typically not approached together in a single model. GeoMELT can replace the usage of much larger LVLMs, which can have a prohibitive cost due to their memory requirements, particularly in edge devices, while gaining abilities for retrieval tasks.

Despite the promising results, multiple avenues of progress remain open. In particular, the development of an efficient architecture capable of handling multisensor data and varying spatial and spectral resolutions, without complex architectures, remains particularly interesting. Regarding other possible tasks to be addressed, image change captioning~\cite{liu2022remote,irvin2025teochat}, and composed image retrieval~\cite{psomas2024composed} would also be important for further development. Composed image retrieval~\cite{psomas2024composed}, in particular, would be a great fit for the cross-encoder architecture of GeoMELT, given the need for modeling queries that involve images and text simultaneously. Also, despite the multiple advantages of GeoMELT over the usage of LVLMs shown in this work, GeoMELT cannot engage in dialogue or multi-turn conversation with a user, contrarily to approaches like GeoChat. However, GeoMELT can still be integrated in pipelines involving LVLMs, e.g., for retrieval augmented generation~\cite{ramos2022smallcap} by retrieving image or text examples for a given prompt, or acting as a scoring function for quality-aware decoding methods. Finally, it remains worth exploring the reasoning capabilities of GeoMELT, which, probably due to its simple design, and as suggested by the VQA results, currently do not match the reasoning abilities of LVLMs.

\section*{Acknowledgments}

This research was supported by the Portuguese Recovery and Resilience Plan through project C645008882-00000055 (i.e., the Center For Responsible AI), and by Fundação para a Ciência e Tecnologia, I.P. (FCT) through the projects with references UID/50021/2025 and UID/PRR/50021/2025.
Devis Tuia was also supported by the Horizon Europe grant 101213369 DVPS.


\bibliographystyle{IEEEtran}
\bibliography{tidy}


 
\begin{IEEEbiography}[{\includegraphics[width=1in,height=1.25in,clip,keepaspectratio]{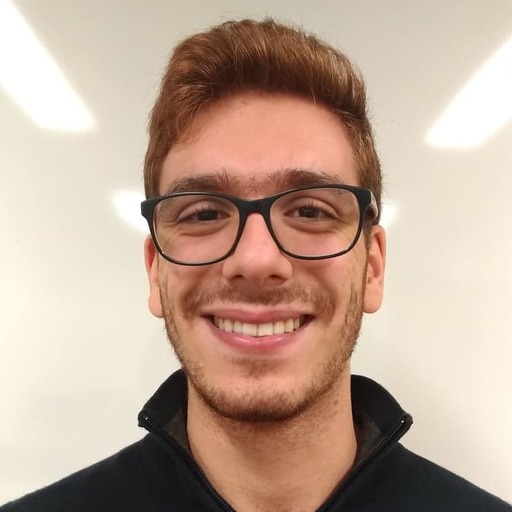}}]{João Daniel Silva}
received the M.Sc. degree in computer science from the Instituto Superior Técnico, University of Lisbon, in 2021, where he is currently pursuing the Ph.D. degree in Electrical and Computer Engineering. He is also an Early Stage Researcher at the Human Language Technologies Laboratory, INESC-ID. His research interests include vision and language tasks, multimodal machine learning, and the analysis of remote sensing images.
\end{IEEEbiography}
\vspace{1pt}

\begin{IEEEbiography}[{\includegraphics[width=1in,height=1.25in,clip,keepaspectratio]{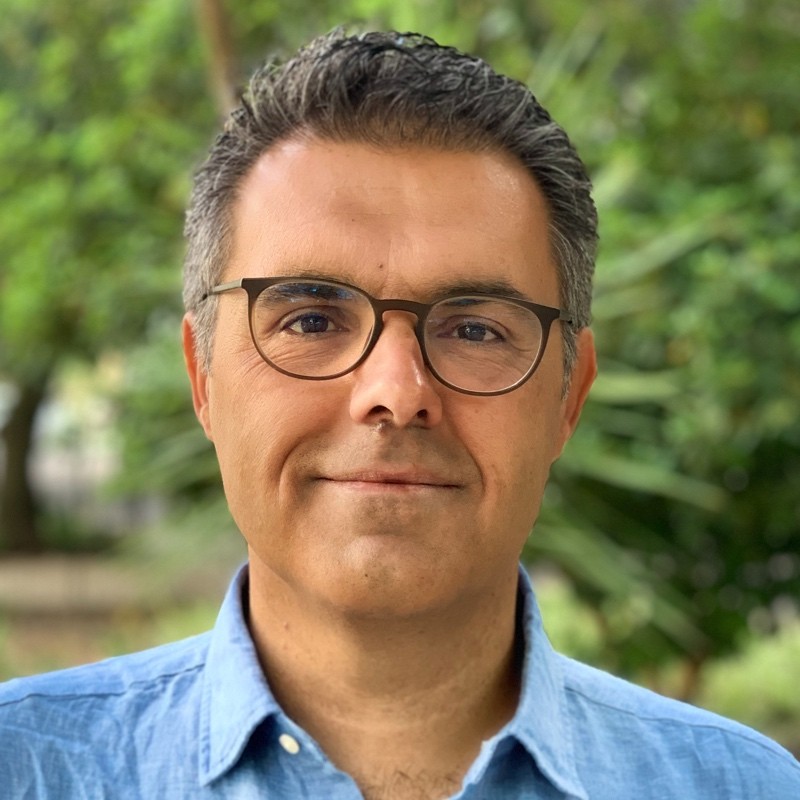}}]{João Magalhães}
João Magalhães (PhD, Imperial College London, UK, 2008) is a Full Professor at the Department of Computer Science at the Universidade NOVA de Lisboa. His research interests cover the different problems of vision and language understanding, in particular: foundational vision and language models, multimedia search, multimodal conversational AI, and multimodal temporal models. He has coordinated and participated in several research projects, where he aims to generalize his vision and language research to solve real-world problems across different domains. The work of his group has been awarded or nominated for several distinctions: in 2023 and 2022 his group was awarded the 1st place and 2nd place respectively in the Amazon Science Alexa TaskBot Challenge, a conversational AI challenge to incorporate multimodal (voice and vision) user data, in 2020 a best paper award at the Portuguese NLP conference (PROPOR), in 2018 two nominations for best paper at ACM Int’l Conference on the Theory of Information Retrieval and at ACM Int’l Conference in Multimedia Retrieval.
\end{IEEEbiography}

\vspace{1pt}
\begin{IEEEbiography}[{\includegraphics[width=1in,height=1.25in,clip,keepaspectratio]{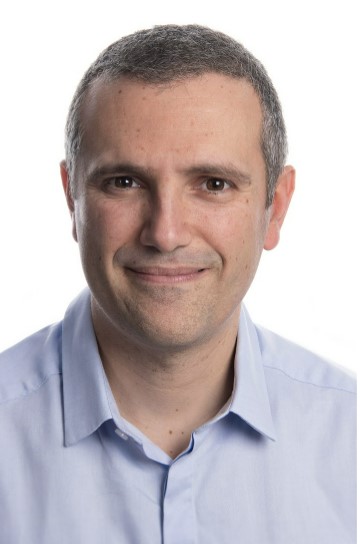}}]{Devis Tuia}
(Senior Member, IEEE) received the Ph.D. degree from the University of Lausanne, Lausanne, Switzerland, in 2009. He was a Post-Doctoral Researcher with Valéncia, Boulder, CO, USA, and École polytechnique fédérale de Lausanne (EPFL), Lausanne. From 2014 to 2017, he was an Assistant Professor with the University of Zurich, Zürich, Switzerland. He then was a Professor at Wageningen University, Wageningen, The Netherlands. Since 2020, he has been an Associate Professor at EPFL-Valais, Sion, Switzerland. His research interests include machine learning and computer vision for spatial data and in particular studying new concepts for AI4EO to make images more accessible and models more understandable.
\end{IEEEbiography}
\begin{IEEEbiography}[{\includegraphics[width=1in,height=1.25in,clip,keepaspectratio]{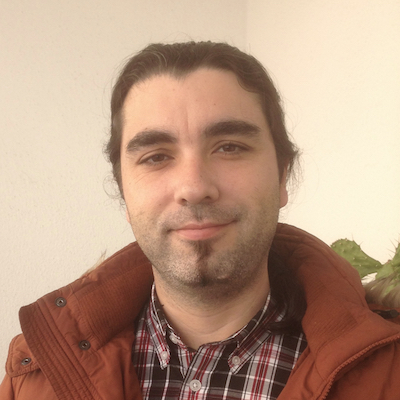}}]{Bruno Martins} (Senior Member, IEEE) was born in Lisbon, Portugal, in 1979. He received the Ph.D. degree in computer science from the University of Lisbon, in 2009.

He is currently an Associate Professor with the Instituto Superior Técnico, University of Lisbon, a Researcher with the Human Language Technologies Laboratory, INESC-ID, and a member of the
Lisbon ELLIS Unit (LUMLIS). He works on problems related to the general areas of information retrieval, text mining, multimodal machine learning, and the geographical information sciences. He has been involved in several research projects related to geospatial aspects in information access and retrieval, and he has accumulated a significant expertise in addressing challenges at the intersection of information retrieval and the geographical information sciences.
\end{IEEEbiography}
\vspace{1pt}



\end{document}